# 2D Path Solutions from a Single Layer Excitable CNN Model

Koray Karahalıloğlu

June 30, 2006

Department of Electrical and Computer Engineering,

Virginia Commonwealth University,

601 West Main St., P.O. Box 843072,

Richmond, VA 23284-3072, USA.

E-mail: kkarahalil@vcu.edu

**Abstract**

An easily implementable path solution algorithm for 2D spatial problems, based on excitable/programmable characteristics of a specific cellular nonlinear network (CNN) model is presented and numerically investigated. The network is a single layer bioinspired model which was also implemented in CMOS technology. It exhibits excitable characteristics with regionally bistable cells. The related response realizes propagations of trigger autowaves, where the excitable mode can be globally preset and reset. It is shown that, obstacle distributions in 2D space can also be directly mapped onto the coupled cell array in the network. Combining these two features, the network model can serve as the main block in a 2D path computing processor. The related algorithm and configurations are numerically experimented with circuit level parameters and performance estimations are also presented. The simplicity of the model also allows alternative technology and device level implementation, which may become critical in autonomous processor design of related micro or nanoscale robotic applications.

# 1 Introduction

The CNN theory can be applied to various collective system dynamics in nature. An interesting example is the realization of excitable dynamics, such as reaction-diffusion system dynamics via autonomous CNNs [1], [2]. Several application examples of such systems include artificial locomotion [3]-[6], image processing [7] and shortest path solution [8]. The spatiotemporal phenomena in existing CNN-UM implementations were also reported recently [9]. Herein we consider a much simpler excitable model, which is a CNN layer inspired by the excitation dynamics in distributed FitzHugh-Nagumo (FHN) neuroelectric model [10]. The distributed (or diffused) variants of FHN model are well-known and have been surveyed in the literature such as in [11], including top-down CMOS implementation such as in [12]. The excitable network considered here preserves the essential functionality required for 2D path computation with autowaves, while remaining as a single layer model. The resistively coupled cells in the network exhibit a cubic polynomial I-V response as the reaction function (Fig. 1). The related cell states correspond to the voltage state of each cell node, where the response is in general bistable. At a critically biased level in the phase space, this network model is capable of high state propagation as trigger autowaves across the single layer 2D cell array. The corresponding excitable characteristic was implemented by custom analog circuit design, which was confirmed with circuit level numerical studies [13] and more recently through CMOS implementation [14].

An additional applied feature regards a 2D spatial configuration of resistive cell coupling in the model. This configurability is analogous to arrangement of a symmetric template of state coupling in CNN terminology. With such arrangements in the 2D cell array, spatial obstacle data can be directly programmed onto the network area. A combination of this feature with the excitable state characteristic, results in a single layer CNN capable of path solutions in 2D space with arbitrary obstacles. Due to this single layer architecture, it is possible to obtain high resolutions in on-chip implementations for such applications.

The presented work focuses on a CMOS design parameters, due to the conventional availability of this technology. However, because of the familiar reaction function behavior, the same model can also be realized at the device level and through emerging technology methods, such as by utilizing resonant tunnel diodes (RTD) or self-assembled nanodevice structures [15]. The characteristic responses of such devices are especially suitable for implementation of the particular reaction function. These additional possibilities also render the model as appealing, especially if one considers the adaptation potential to nanoscale robotic applications where autonomous path optimization is required. The related application of the model is therefore investigated numerically, with path solution examples employing CMOS circuit parameters and a typical array size achievable in this technology. The overall performance estimations, with solution times for the given examples are also presented.

## 2 Excitable Network

The resistive-coupled single layer CNN model in Fig. 1 can serve as an excitable wave propagating network. For each cell $ij$ in the 2D array, the cell state equation is expressed as

$$\dot{v}_{ij} = \frac{G}{C_I}(v_{i+1,j} + v_{i-1,j} + v_{i,j-1} + v_{i,j+1} - 4v_{ij}) + \frac{I_B - J_{ij}(v_{ij})}{C_I} \qquad (1)$$

Here, node voltage $v_{ij}$ corresponds to the cell voltage state, $I_B$ is a common bias current for each cell, and $J_{ij}(v_{ij})$ is the nonlinear current response of the cell $ij$. The capacitance $C_I$ can either represent an external capacitance or an approximation to the internal cell circuit capacitance.

The linear coupling of the cell states $v_{ij}$ in equation (1) form a discrete space Laplacian. Therefore, the diffusion dynamics is already inherent. Several functions can be candidate reaction terms, as classified for reaction-diffusion models such as in [16]. Herein we consider a bistable variant in parallel with the case of FHN model. In the related excitation dynamics, the nonlinear reaction function is cubic (polynomial) in characteristic, as a function of the state variable. Therefore, provided that $I_B - J_{ij}(v_{ij})$ realizes a similar reaction term in (1), this equation will exhibit an essential part of the reaction-diffusion dynamics in the FHN model, providing an excitable wave propagating response.

An approximate view of the single cell phase space can be given as in Fig. 1, which also analytically holds if one decouples the neighbor cell contribution in (1) approximated as an independent function

$$g(t) = \frac{G}{C_I}(v_{i+1,j} + v_{i-1,j} + v_{i,j-1} + v_{i,j+1}) \quad (2)$$

The real cell coupling is bidirectional however, this assumption shows the main state dynamics in the phase space, since one can write (1) as

$$\dot{v}_{ij} \approx -\frac{4G}{C_I}v_{ij} + g(t) + \frac{I_B - J_{ij}(v_{ij})}{C_I} \quad (3)$$

Therefore, if a cubic I-V response $J_{ij}(v_{ij})$ from the cells and an offset bias current $I_B$ are established, a bistable condition in the phase space can be achieved (Fig. 1). The specific reaction function is of particular interest, since it provides this bistable condition and the cubic characteristic is already available, through the device level implementation options.

## 3 Stable Cell States

The stable cell states can easily be expressed in the network regions where all the cells assume identical stable states, as low state $V_L$ or high state $V_H$. Since the states are identical, the discrete Laplacian in (1) vanishes in those regions. A state equilibrium then requires $I_B - J_{ij}(v_{ij}) = 0$. Therefore, one has

$$J_{ijL}(V_L) = I_B; J_{ijH}(V_H) = I_B \qquad (4)$$

where $J_{ijL}$ and $J_{ijH}$ correspond to the low and high state values depending on the cubic function. Given an analytical model for this function, one can determine the *regional* stable states $V_L$ and $V_H$. For instance, a related result was in [14] for the particular CMOS design.

The phase space in Fig. 1 also depicts that an increase in the common bias current $I_B$ shifts the stable states and moves the state curve in positive $\dot{v}_{ij}$ direction for each cell. In addition, for $I_B$ close to the peak current of the I-V response, the low stable point $V_L$ is about to become a saddle node. At this critical stage, the network is in an excitable mode. The propagating state transitions can be introduced by an initial excitation of any cell state. After a saddle node transition, the only stable state for the excited cell is $V_H$. This transition causes a domino effect leading to successive coupled cell transitions and establishing a high state propagation across the network as a trigger wave. This propagation is different from a merely diffusion based response, and the waves are non-dissipative. On the other hand, without the initial excitation, all the cell states still remain at excitable low stable states $V_L$.

## 4  Representation of Obstacle Data

The resistive coupling of the network cells correspond to a symmetric CNN template describing the neighbor cell state interactions in CNN theory. If these couplings are arranged with particular spatial distributions, the autowave propagation in the specified regions of the network area can be prevented. Therefore, the resistive coupling of cells provides direct means to represent the spatial distribution of obstacles. It is possible to store data of such 2D navigational constraints, with direct mapping of the obstacles to cell coupling strength. The mapping of these obstacles can also be updated dynamically, as the navigation occurs. Therefore, both static and dynamical obstacle distribution problems can be addressed.

By the characteristic of discrete propagation dynamics, if the coupling conductances are decreased, this weakened state coupling becomes insufficient for the saddle node transitions triggered by the neighbor cells. Therefore, the propagation fails along specified regions. As a result, trigger waves can be guided across the network area, avoiding regions identified as obstacles. Thus, we set these "forbidden" regions with arrangements of the resistive coupling distribution, representing the obstacle boundaries in a particular path solution problem.

For the related configurations, two static 2D obstacle distribution problems are considered as a *room* with relatively scarce obstacles and a *maze*, both as mapped directly to resistive cell coupling of a 2D cell array. For the programming of related coupling for numerical purpose, an efficient configuration method was previously introduced as the *template image method* [17]. In this method, one defines the coupling of cells based on the pixel intensity information of grey-scale template images. According to particular update rule employed, the coupling resistance between two adjacent cells was assumed as proportional to the absolute value of related pixel intensity difference in the template image. Fig. 2 shows the used template images for the particular simulated examples.

Regarding a hardware implementation approach of this method, in the past row/column addressing was used to program initial conditions to the network cells [14]. This design employed pass transistors operating in resistive mode, which effectively approximate the constant coupling conductance $G$ in (1). However, a more efficient optical sensor based cell coupling can be realized for fast programming of obstacle data. Such an integration can provide the related input data in real-time, as a bird's eye view of the navigated area. A similar sensor integration for state programming of the cells was already implemented [18].

## 5  Path Solution Algorithm

Based on an *absolute* coordinate mapping of 2D spatial data for a given problem, a simple path solution algorithm can be applied via the combination of excitable and programmable network characteristics: An estimated direction or an exact target coordinate (TC) is specified as the source of autowaves via related cell coordinate addressing and an additional current excitation of this cell during excitable mode. This temporary excitation starts an autowave originating from the target cell, propagating around the programmed obstacle regions, as the wave propagation fails across obstacle boundaries due to cell decoupling. In addition, a reference point can be defined in order to indicate the current coordinates of the robotic vehicle, which is to be updated according to the autowave propagations. In general, there is more than one available path from target to a given reference coordinate, however the constant speed wavefront travelling the *shortest path* will first arrive at a neighbor cell of the reference cell coordinate, thus triggering its state change. Here we use the term *shortest path* strictly for the 4-neighborhood connectivity model. Since a diagonal neighborhood is not defined, the *physical* shortest path in 2D discrete space is not computed. In accordance with above, the first detected state transition among the neighbor cells signals the correct direction towards the target. Then, the reference cell coordinate (RC) for the next iteration is set to this *winner* cell coordinate. The network biasing $I_B$ is removed to eliminate further propagations, which also resets the cell states to $V_L$ for the next iteration cycle. After the update of the reference cell coordinate, another wave is propagated from the target coordinate and neighbor cells of the new reference cell are scanned for state transitions. The iterations continue until the target cell coordinates may be reached. The successively stored *winner* cells upon iterations provide a path solution to the 2D problem mapped onto the network. In the case there are no available paths between the reference and target coordinates, the autowave will not reach any neighbor cells of the reference cell. Therefore, a no solution case can also be determined in a single iteration cycle and through related timeout criteria. The above steps are summarized as a solution algorithm in Fig 3. In general, this solution algorithm will remain valid without obstacle or target related requirements, such as their shapes, change or motion. This argument can hold as long as a sufficient array resolution is available for the given problem and the autowave speed is faster compared to the rate of change in obstacle related spatial data. For problems involving dynamical changes, cell coupling has to be updated before each new iteration, introducing the current locations of obstacles. Also, the wave propagation is then triggered from the most recent target coordinates.

The computational performance is governed by the network size (spatial resolution), complexity of the obstacle data and system parameters determining the propagation speed of trigger waves. A related important effect regards the common bias current $I_B$. Within the bistable range in the phase space, a high value of $I_B$ leads to accelerated saddle node transitions, resulting in faster wave propagation. However, this bias current should not exceed the peak current value of the cubic function, where these transitions will take place without excitation. Therefore, the optimized performance depends on how accurately the bias current is controlled in a practical design, also by overcoming noise and device variations. On the other hand, due to the simple architecture, the peripheral access and control tasks can be kept to a minimum. The essential cell access task involves a periodic row/column addressed scanning of four neighbor cell states around the current reference cell, which is to be updated upon the *winner* cell detection. The generated autowave or each cell in the array need not be monitored, therefore for large arrays, the *winner* cell detection and reference cell update times are short, compared to the autowave propagation time in each iteration. As the most significant component of the computational performance becomes the autowave propagation speed, a related analytical approach is very desirable for design. In the described network model however, an analytical solution of propagation speed is not straightforward. The state equation (1) describing the dynamics is a nonlinear differential-difference equation. There are studies for similar excitable dynamics with simplified activating functions in literature, in continuous space such as [19] or discrete space as in [20], [21]. The analytical methods in general allow simplified reaction functions mostly providing implicit relations for the propagation speed. As this nonlinear function form is very definitive, its first order representation can lead to considerable deviations in analytical estimations. Hence, we limit the current analysis with numerical results, which incorporate the previously implemented circuit parameters in order to provide implementation based performance estimations.

## 6 Numerical Experiments

The network dynamics and path solution algorithm are experimented numerically, in order to validate the proposed functionality and performance based on the circuit parameters of a previous hardware implementation. For this purpose, a dedicated analysis tool is used where the described algorithm is also incorporated. The circuit model and a piecewise nonlinear approximation of the I-V response from a previously implemented MOS cell circuit is employed in the simulations [14], as shown in Fig. 4. Here $J_P$ denotes the peak current from this response. The used parameter values in the examples are $I_B = 21\mu A$, $G = 25\mu S$ (if coupled), $C_I = 500 fF$. These values result in $V_H = 1.75V$ and $V_L = 0.97V$ as regional stable states of the excitable network. In accordance with this range, the voltage range 0-1.8 *V* is linearly mapped to grey-scale pixel intensity for visualization of the voltage state changes and overall system dynamics. The template images (Fig. 2) are utilized to configure the 2D cell array within the numerical tool. The contrasting adjacent pixel regions are therefore effectively decoupled, which also identify the obstacle regions. The rest of the coupling conductances are constant throughout the network. As a result, the excited autowaves fail along the spatial boundaries of these obstacles. The simulated circuit level dynamical response for the *room* example is shown in Fig. 5 and for the *maze* example in Fig. 6.

Together with the coupling configurations, the described algorithm is also incorporated with the analysis tool. Therefore, together with the dynamical response, a *winner* cell is evaluated upon each wave propagation, implementing the solution algorithm. The *winner* cell criteria is applied as the first neighbor cell achieving the state level $V_W = 1.2V$, upon each iteration. A more general criteria can be the definition of a *winner* threshold voltage $V_W$, around $V_W \approx (V_L + V_H)/2$. This level during state transitions is more accurately detectable, due to the almost linear time response from the cell dynamics. The solution examples reaching to the target (wave source) from different start points with number of steps required (or the path length) are given in Fig. 7 and Fig. 8 for the *room* and *maze* examples respectively. The simulated examples specify an initial reference (or start) point (R.pt.), the fixed target point to be reached (T.pt.) and arrows indicating the direction of solution steps. The paths shown are constructed by combination of the *winner* cells upon wave propagation in each iteration. The simulated state transitions in time domain as a trigger wave are also presented in Fig. 9. Here, the time interval between transitions is defined by $t_P$, with respect to voltage level $V_W$. The numerical result obtained via used parameters was $t_P = 16.92\,ns$. Accordingly, the propagation speed becomes $c_p = 1/t_P = 59.1\,cells/\mu s$. Thus without obstacles, the completion of trigger wave propagation is expected around 2 $\mu s$ for the particular array size and parameters. The propagation speed variations with changing bias current $I_B$ are also experimented and Table 1 depicts the related numerical results. A significant variation is observed within a narrow range of $I_B$ values, where all other parameters are kept as fixed. Although higher speed is achievable for large $I_B (<J_P)$ values, preserving the bistable condition simultaneously will require a precise control of this current, as very close to $J_P$. Thus, a related optimization is critical in designs for this particular application.

Comparing the number of solution steps in Fig. 7 and Fig. 8, the spatial coverage duration of a constant speed wave also depends on complexity of the obstacle distributions, hence the particular path problem. This for instance, renders the iterative steps longer for the *maze* example. Herein, we present a simple analysis which can estimate the time performance of the model for large arrays. Especially for such high resolution applications, the *winner* cell detection, update and storage times are expected to be an order of magnitude lower compared to autowave travel durations across the network area. Therefore essentially, the autowave travel time upon each iteration can be considered as the most significant contributor. For a fixed point target example, the total solution time can be deduced easily. Let $P$ indicate the path length as number of cells (or steps) in the solution path, excluding the reference cell. Starting from the target cell and by above definition, there are ($P$-1) state transitions in time, until a neighbor cell of the reference cell is reached and related state transition can be detected. After the first iteration and update of the reference cell, there remains ($P$-2) and so on. Hence, neglecting the related detection and update times, total time required for a solution with $P$ steps follows from a series summation as

$$T_S = \sum_{n=1}^{P-1} n \times t_p = P(P-1)\frac{t_p}{2} \qquad (5)$$

Based on this result, the longest durations of solutions are given as for instance, $T_S = 164.63\ \mu s$ (140 steps) for the *room* problem, and $T_S = 465.21\ \mu s$ (235 steps) for the *maze* problem, also by referring to Fig. 7 and Fig. 8 respectively. It should be noted that, these solution steps can be computed on-the-run as well. Hence, these total durations do not correspond to the response time of a robotic vehicle. However as (5) dependency implies, one of the main issues in performance regards a reduction of solution steps $P$, hence the array resolution reserved for the definition of a given problem, whenever possible.

A worst case assumption can also follow if a maximum path length of $P_{max} \approx (N \times M)/2$ is considered for an $N \times M$ 2D array size. This is a reasonable upper limit for the solution path length, which may still allow autowave propagation between two distant points in the 2D array. Substituting this $P_{max}$ into (5) gives a maximum solution time as

$$T_{S\max} = (N \times M)((N \times M)/2 - 1)\frac{t_p}{4} \qquad (6)$$

According to the worst case scenario, the particular 80×80 array size and used circuit parameters will require $T_{S\max} = 86.60\ ms$ to complete such a solution. If one considers the complexity of general problems, this can amount to a significant real-time performance, as the processor provides the solution steps on the run. In addition, note that $t_P$ is a direct function of the cell capacitance $C_I$ in the model. In a previous hardware implementation, an external capacitor (500 *fF*) was maintained for analog storage of initial voltage states for the cells, regarding various other applications. The same order of magnitude is employed in the presented numerical experiments as a conservative estimation. Nevertheless, a high capacitance is not required for this particular application, therefore up to a certain level, the given propagation speeds can be improved by design as well. This is especially valid for possible alternative technology design of the same model.

The numerical results also verify that although a diagonal path length optimization is not possible, a close approximating response is achieved. Such a response is expected from a 4-neighborhood connectivity model and its related implementations as well. If diagonal cell coupling is incorporated, a closer path length optimization can be realized. However in that case, the implementation complexity is increased. It is also worthwhile to note that compared to trigger waves, front waves available from various multilayer excitable CNN dynamics can significantly reduce the solution times. Then the solution algorithm can be performed between the successive wavefronts with a sufficiently large "wavelength" in discrete cell space, to allow completion of *winner* cell evaluation process. Such a multilayer approach on the other hand, again comes with compounded layout area overhead in implementation. Hence, especially when compact functionality is of primary importance, this single layer CNN model can provide a better alternative in related processor design.

## 7  Conclusion

2D path solutions which follow from the excitable dynamics of a single layer CNN model are numerically investigated, with example results of the employed solution algorithm. The method can be applied to general navigation problems as defined in 2D space via direct mapping of obstacle distributions to resistive cell coupling in the 2D cell array. The numerical experiments are based on circuit parameters of a previous CMOS implementation. The results indicate that the network dynamics can indeed perform path solutions along arbitrary obstacle distributions. In addition, the generated solutions give close approximations to optimum path lengths. The propagation speed is the key element of related computational performance. Accordingly, certain computational performance estimations are presented. The results point to a significant propagation speed achievable with rather conservative parameters. They also indicate that a precisely controlled large bias current in the related design will be very significant for the autowave speeds. Another important parameter is the cell capacitance in the model. These two contributions especially need to be addressed in order to optimize the wave speed. The main advantage of the excitable model comes from its single layer simple architecture. If employed as the core section of a path solver processor, the model can allow layout flexibility for various design and interfacing techniques, such as different sensor integration methods, as well as emerging technology and device level implementations. These alternatives can become particularly important in processor designs for micro and nanoscale robotic applications. A higher spatial resolution is also feasible via a single layer CNN, compared to a multilayer excitable model. A typical array size of 100×100 cells is estimated via conventional submicron CMOS process, which is larger than the array size used in the presented numerical examples.

Table 1: Numerical results regarding propagation speed variation with bias current $I_B$.

| $I_B(\mu A)$ | 18 | 19 | 20 | 21 | 22 | 23 |
|---|---|---|---|---|---|---|
| $t_p(ns)$ | 36.06 | 27.02 | 21.19 | 16.92 | 13.43 | 10.22 |
| $c_p(cells/\mu s)$ | 27.7 | 37.0 | 47.2 | 59.1 | 74.4 | 97.8 |

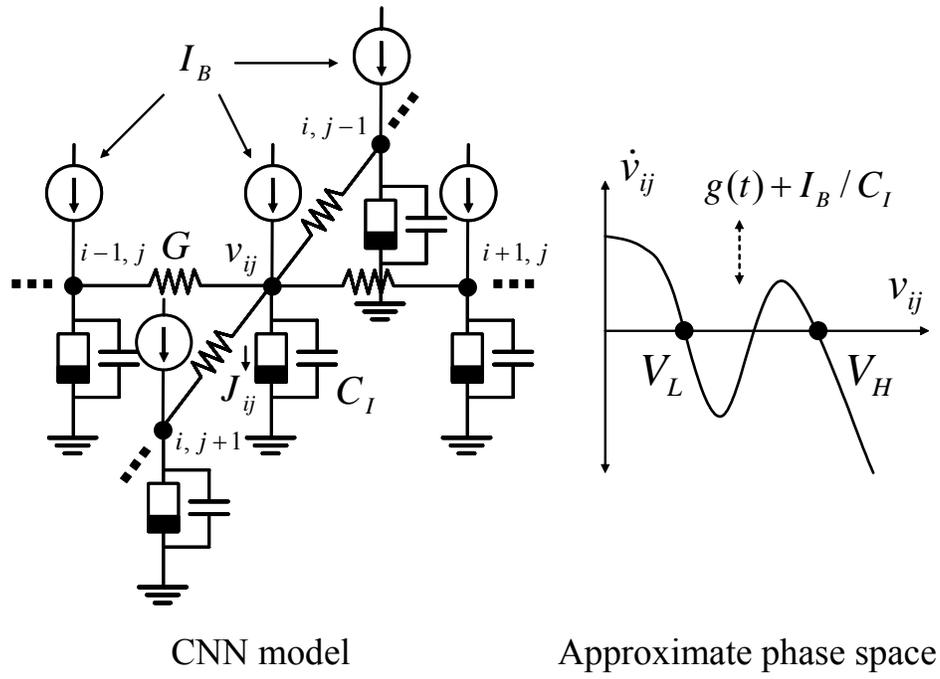

Figure 1: An excitable single layer CNN model with approximate phase space for each cell *ij*.

Template Images Providing the Obstacle Data

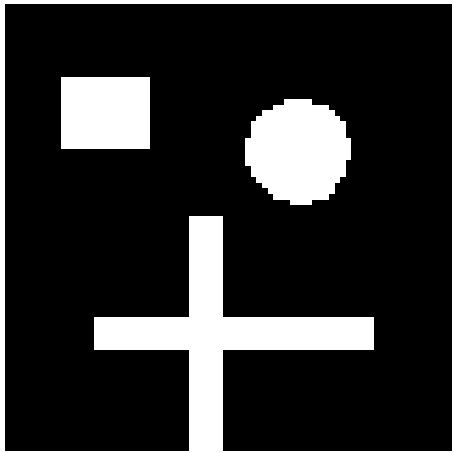 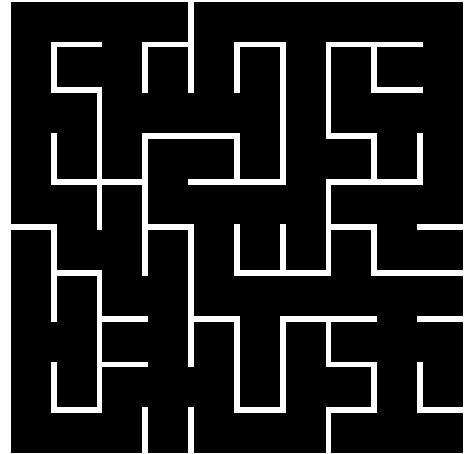

*Room*   *Maze*

Figure 2: Template images (80×80) employed in numerical analysis for configuration of resistive cell coupling, for two examples of 2D obstacle data

1. Set/update the target coordinate (TC)

2. Set/update reference coordinate (RC)
   (RC = last *winner* cell coordinate)

3. If TC = RC target is reached, end iterations

4. Map the current obstacle data to cell coupling

5. Bias the network to excitable state mode

6. Stimulate autowave propagation from TC

7. Detect/store the *winner* cell coordinate

8. Reset all cell states to low

9. Go to step 1

Figure 3: Iterative steps of the path solution algorithm

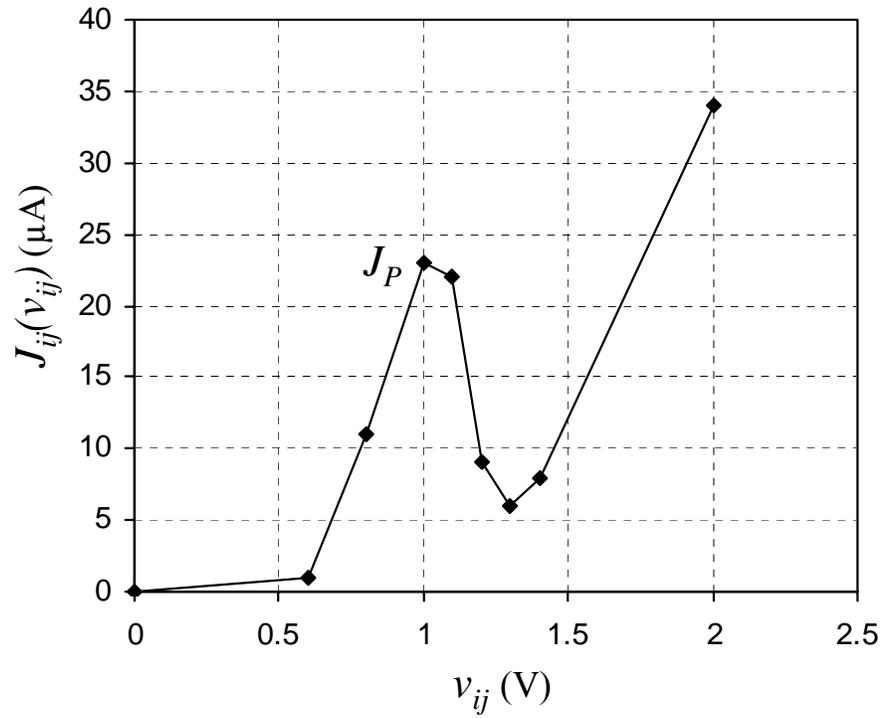

Figure 4: Piecewise nonlinear approximation of the measured I-V response from a previously implemented MOS cell circuit, employed in large cell array numerical analysis

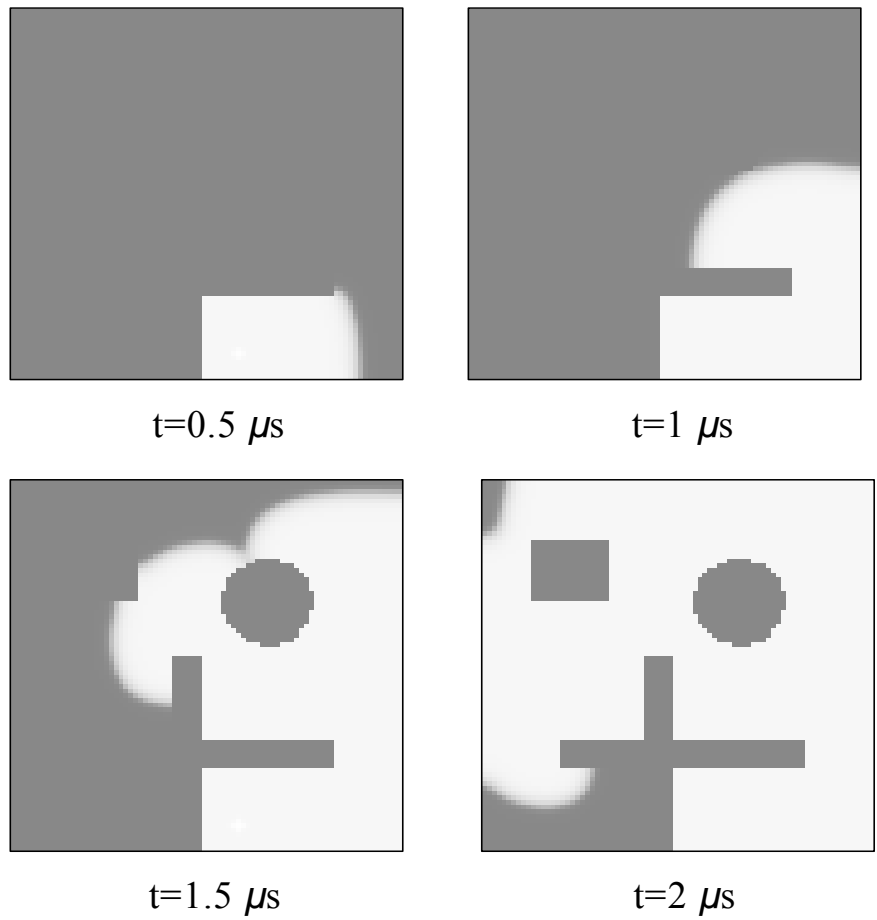

Figure 5: *Room* example: Trigger wave propagation of voltage states from 80×80 array circuit level simulation

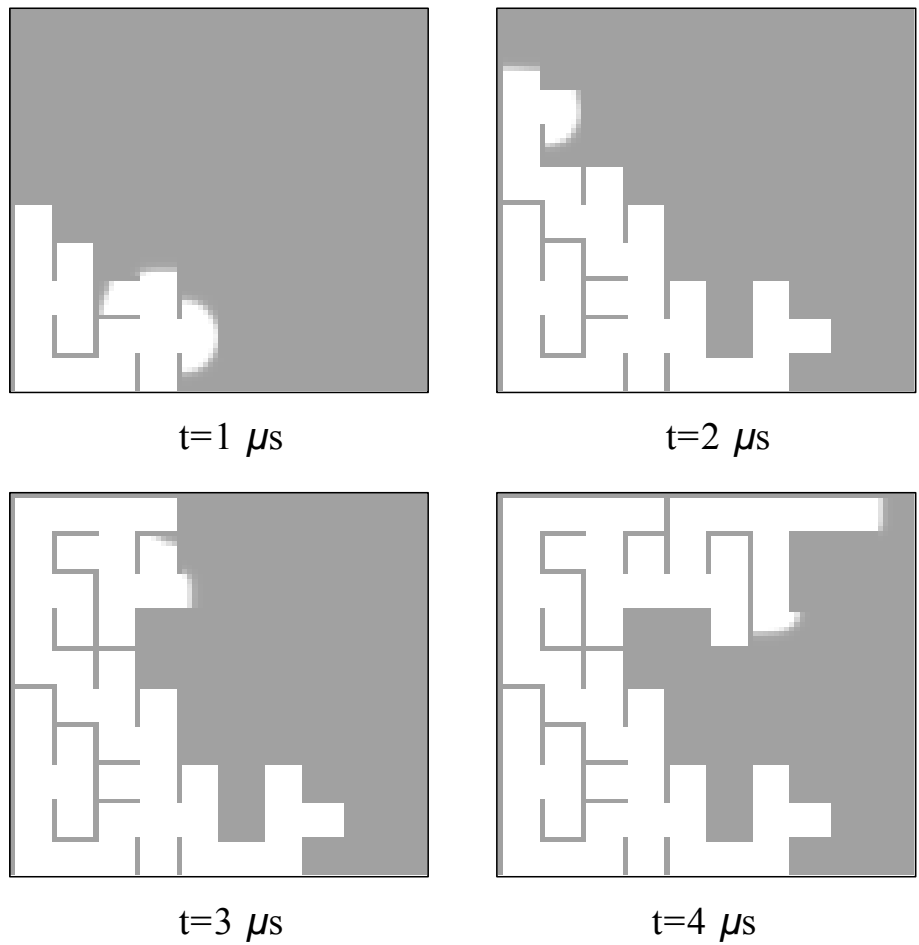

Figure 6: *Maze* example: Trigger wave propagation of voltage states from 80×80 array circuit level simulation

Figure 7: Various solutions and their number of iterations across the *room* from different initial reference points (R.pt.) towards the same target point (T.pt.)

Figure 8: Various solutions and their number of iterations through the *maze* from different initial reference points (R.pt.) towards the same target point (T.pt.)

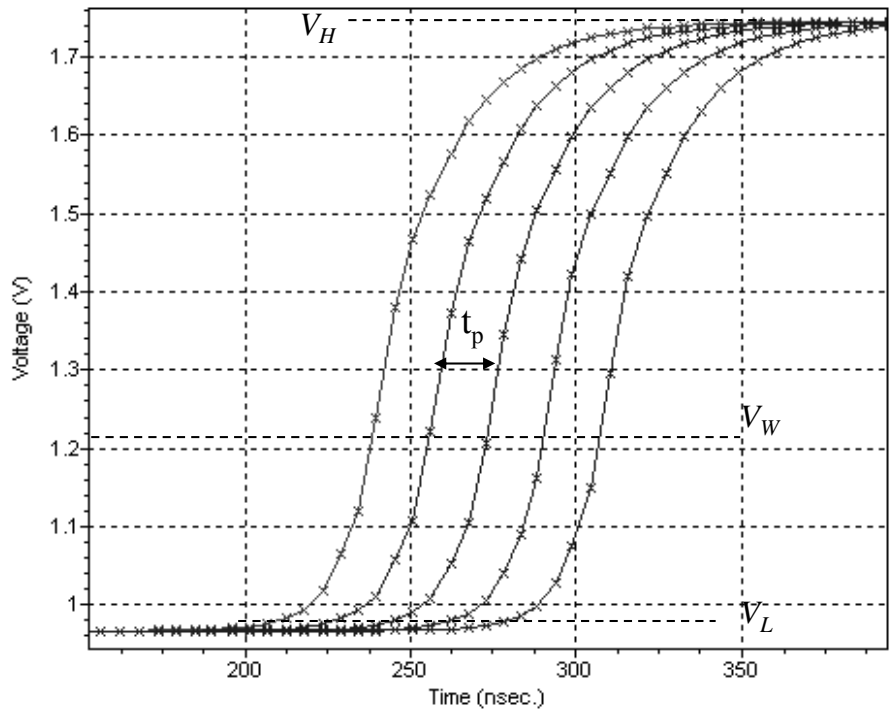

Figure 9: Successive triggered state transitions from a section of the 2D cell array, propagating as a trigger wave